\documentclass{article} 
\usepackage{iclr2027_arxiv,times}

\usepackage{amsmath,amsfonts,bm}

\def\eqref#1{equation~\ref{#1}}
\def\Eqref#1{Equation~\ref{#1}}

\def\1{\bm{1}}

\DeclareMathAlphabet{\mathsfit}{\encodingdefault}{\sfdefault}{m}{sl}
\SetMathAlphabet{\mathsfit}{bold}{\encodingdefault}{\sfdefault}{bx}{n}

\usepackage{amsmath,amssymb,amsthm}
\usepackage{booktabs}
\usepackage{graphicx}
\usepackage{xcolor}
\usepackage{tikz}
\usepackage{hyperref}
\usepackage{url}
\usepackage{natbib}
\usetikzlibrary{arrows.meta,positioning,shapes.geometric,fit}
\usepackage{algorithm}
\usepackage{algpseudocode}
\usepackage{amsmath}
\usepackage{amssymb}
\usepackage{booktabs}
\usepackage{graphicx}
\usepackage{todonotes}
\presetkeys{todonotes}{inline}{}

\usepackage{longtable}
\usepackage{fancyvrb}
\usepackage{titletoc}

\definecolor{recountblue}{HTML}{DCEAF7}
\definecolor{recountblueedge}{HTML}{4C78A8}
\definecolor{recountorange}{HTML}{FBE6D2}
\definecolor{recountorangeedge}{HTML}{D97832}
\definecolor{recountgreen}{HTML}{DDEFE1}
\definecolor{recountgreenedge}{HTML}{4E8B57}
\definecolor{recountgray}{HTML}{EEF0F2}

\title{Denoising Blocks, Not Tokens: Efficient Compressed Continuous Diffusion with Branching Token Realization}

\author{
Xinsong Feng$^{1}$\thanks{Work done during an internship with the Writer AI Research team.},
Peng Du$^{2}$,
Zhizhuo Yang$^{2}$,
Daniel M. Bikel$^{2}$,
Jiayun Wang$^{1}$,
Haipeng Chen$^{3}$
\\
$^{1}$Georgia Institute of Technology,
$^{2}$Writer AI Research,
$^{3}$William \& Mary
}

\iclrfinalcopy 
\begin{document}

\maketitle

\begin{abstract}

Diffusion language models (DLMs) generate text through iterative parallel refinement, offering the potential for higher throughput than autoregressive (AR) decoding.
However, most DLMs still maintain one generative state per token, so every denoising step processes a state sequence as long as the output sequence, limiting the throughput gains from parallel generation.
Continuous DLMs provide an additional degree of freedom: a single continuous state can represent multiple tokens, allowing diffusion to operate on a much shorter latent sequence.
We introduce \emph{Branching Latent Diffusion (BLD)}, which exploits this flexibility by compressing a 1024-token sequence into only 64 block latents, a $16\times$ reduction.
BLD combines latent compression with \emph{branching token realization}, where each latent is decoded by a local AR branch and all branches run in parallel.
Because strong compression makes joint latent generation difficult, BLD generates the latents in groups, conditioning each group on previously generated latents.
In end-to-end evaluation on the same GPU, BLD reduces generation FLOPs by more than $80\times$ and increases throughput by more than $6\times$ relative to the similarly sized ELF-L baseline.
Compared with the AR baseline, BLD achieves more than $6\times$ higher throughput and more than $4\times$ lower latency.
Despite the compression, BLD maintains competitive local fluency and diversity, although long-range coherence remains challenging.
Overall, BLD shows that moving diffusion from token-level states to compressed latent sequences can substantially improve the efficiency of long-sequence generation.

\end{abstract}

\section{Introduction}


Autoregressive (AR) language models generate text one token at a time, making generation inherently sequential.
Diffusion language models (DLMs) relax this constraint through iterative parallel generation, and recent advances in discrete diffusion and masked generation have substantially improved generation quality and sampling efficiency~\citep{lou2024discrete,sahoo2024simple,arriola2025block,nie2025large}.
However, parallel generation reduces sequential steps without reducing the sequence length processed at each step: DLMs typically maintain one state per token, so every denoising step still operates over the full output length.
While this granularity is intrinsic to discrete token states, continuous states need not correspond one-to-one with tokens.
Coarsening continuous generative states therefore offers a direct way to reduce generation cost.

Continuous DLMs provide a natural opportunity to reduce this cost~\citep{li2022diffusion,lin2023text,gao2024empowering,shabalin2025tencdm,chen2026langflow,hu2026elf,yang2026continuous}.
A continuous generative state need not correspond to a single output token and can instead represent multiple tokens.
This flexibility suggests a promising route to more efficient diffusion: compress a sequence of $L$ tokens into $K \ll L$ continuous states and perform diffusion over the compressed latent sequence.

Compression alone, however, is insufficient for end-to-end efficiency.
If the $K$ generated latents are converted back to text through an $L$-step autoregressive decoder, the sequential bottleneck is merely shifted from latent generation to token realization.
Efficient latent compression must therefore be paired with a decoding mechanism that also avoids token-level sequential generation.

Meanwhile, strong compression introduces another challenge.
As fewer latent states represent the same sequence, each state must encode more information, making the joint latent distribution harder to model.
In our experiments, fully parallel diffusion over strongly compressed latents preserves local fluency but struggles with long-range coherence.
Thus, compression shortens the diffusion sequence but makes latent generation more difficult.

These observations lead to the central research question:
\emph{Can continuous language diffusion exploit strong latent compression while remaining efficient at both latent generation and token realization?}

{\bf Our approach}. We address this question with \emph{Branching Latent Diffusion (BLD)}.
BLD compresses a token sequence into a much shorter sequence of block latents and performs diffusion in the compressed space.
BLD combines this compression with \emph{branching token realization}, where each latent represents a block of tokens and is decoded by a local autoregressive branch, with all branches running in parallel.
As a result, the denoiser operates on only $K$ latent states, while the sequential depth of token decoding is reduced from $L$ to the length of a local branch, approximately $L/K$.

The remaining challenge is to generate the compressed latent sequence reliably.
Rather than denoising all $K$ latents simultaneously, BLD generates the latents in groups of $G$ blocks.
Latents within each group are denoised jointly, while previously generated groups provide clean conditioning context.
This factorization replaces a single difficult global generation problem with a sequence of smaller conditional diffusion problems while retaining parallelism within each group.
The group size controls the trade-off between parallelism and modeling difficulty: $G=K$ recovers fully parallel latent diffusion, whereas $G=1$ generates one latent block at a time.

In our configuration, BLD uses $16\times$ token-to-latent compression, reducing a 1024-token sequence to 64 block latents.
On the same GPU, BLD reduces generation FLOPs by more than $80\times$ and increases throughput by more than $6\times$ relative to the similarly sized ELF-L baseline~\citep{hu2026elf}.
Compared with the AR baseline, BLD achieves more than $6\times$ higher throughput and more than $4\times$ lower latency.
Its single-sequence latency is slightly higher than ELF-L because grouped sampling introduces additional sequential depth at the latent level.
Despite the compression, BLD maintains competitive generation quality, although long-range coherence remains challenging.
These results show that compressed latent diffusion can substantially improve long-sequence generation efficiency when token realization is parallelized.

Our main contributions are threefold:
(i) we introduce BLD, a continuous latent diffusion framework that strongly compresses the generative sequence and translates this compression into end-to-end generation efficiency;
(ii) we realize efficient generation in this compressed space through \emph{branching token realization} and grouped latent generation, reducing token-level sequential depth while mitigating the difficulty of generating all compressed latents jointly; and
(iii) we demonstrate substantial improvements in generation FLOPs and batch throughput over size-matched continuous DLM and AR baselines, and in single-sequence latency over the AR baseline, while analyzing the remaining challenge of long-range coherence.

\section{Branching Latent Diffusion}

\begin{figure*}[t]
\centering
\includegraphics[width=0.9\textwidth]{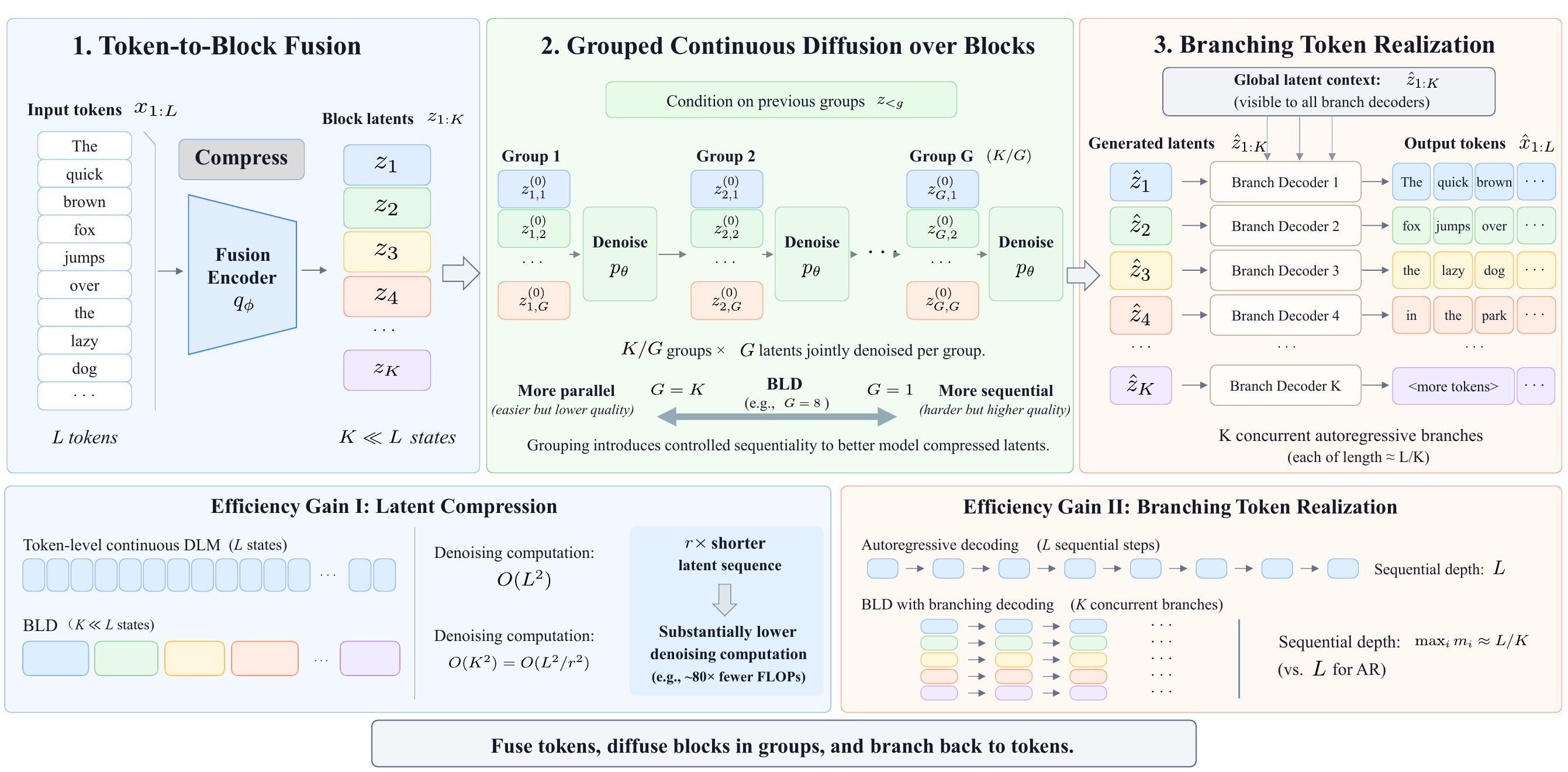}
\caption{
Overview of BLD.
A fusion encoder compresses $L$ tokens into $K \ll L$ block latents.
The latent sequence is partitioned into groups of $G$ blocks and generated autoregressively across groups, with each group conditioned on the previously generated latents.
A branching decoder then expands the complete latent sequence into $K$ token spans and decodes them in parallel.
}
\label{fig:bld_overview}
\end{figure*}

In this section, we present Branching Latent Diffusion (BLD), a continuous latent diffusion framework for language generation that combines compressed latent modeling with parallel branching token realization.
As illustrated in Figure~\ref{fig:bld_overview}, BLD consists of three stages: token-to-block fusion, grouped continuous diffusion over block latents, and branching token realization.
Token-to-block fusion compresses the input sequence into block latents, grouped latent diffusion generates these latents autoregressively across groups, and branching token realization maps the generated latents back to token spans in parallel.
BLD performs global modeling over the compressed latent sequence, while restricting autoregressive token generation to short within-block spans.

\subsection{Token-to-Block Fusion}

Given a token sequence $x_{1:L}\in\mathcal{V}^L$, the fusion encoder deterministically maps it to a compressed latent representation
\begin{equation}
z_{1:K}=q_\phi(x_{1:L}),
\qquad \forall i\in \{1,\ldots,K\}:\
z_i\in\mathbb{R}^d,
\label{eq:fusion}
\end{equation}
where each block latent $z_i$ corresponds to a contiguous token span.
We define the token-to-block compression ratio as $r=L/K$.
Our default configuration uses $L=1024$ and $K=64$, corresponding to a $16\times$ reduction in sequence length.
To support parallel decoding of token spans, we design the block latents to capture both local content and global context for consistency across blocks.

We first encode the input sequence with a bidirectional Transformer to obtain contextualized token representations,
\begin{equation}
h_{1:L}
=
E_\phi(x_{1:L}),
\qquad
\forall t\in \{1,\ldots,L\}:\ h_t\in\mathbb{R}^{d_h}.
\end{equation}
Because encoding is performed before compression, each token representation can incorporate information from the entire sequence.

The contextualized token representations are then partitioned into $K$ contiguous blocks using dynamic chunking~\citep{hwang2026dynamic}.
The chunker predicts boundary scores from adjacent token representations and selects $K-1$ internal boundaries within approximately uniform windows, yielding exactly $K$ contiguous spans.
For the trained model studied here, these boundaries are nearly uniform on full-length documents (Appendix~\ref{app:block_lengths}).
Let $0=b_0<b_1<\cdots<b_K=L$ denote the resulting block boundaries, with $\mathcal B_i=\{b_{i-1}+1,\ldots,b_i\}$ and span length $m_i=|\mathcal B_i|$.

The token representations within each block are aggregated with learned attention pooling to obtain a single block latent,
\begin{equation}
z_i
=
\operatorname{AttnPool}
\left(
\{h_t \mid t\in\mathcal B_i\}
\right).
\end{equation}
Each block latent is aligned with a contiguous token span while incorporating contextual information from the full sequence.
Together, the contextual encoder, dynamic chunking, and attention pooling instantiate the fusion mapping $q_\phi$ in ~\Eqref{eq:fusion}.
The resulting sequence $z_{1:K}$ provides a compact representation of the input using one latent per block.

\subsection{Grouped Latent Flow Prior}

After obtaining the block latents, we next model their distribution with a latent diffusion prior.
A natural idea is to generate the entire compressed latent sequence in parallel,
\begin{equation}
p_\theta(z_{1:K}),
\end{equation}
by jointly denoising all $K$ block latents.
This fully exploits the shorter latent sequence and provides the highest degree of latent-level parallelism.
However, compression concentrates the information needed to represent the same token sequence into fewer latent states.
This reduces the sequence length without necessarily simplifying the joint structure that the prior must generate from noise.
Empirically (Table~\ref{tab:main} and Appendix~\ref{app:conditional_structure}), we find that fully parallel diffusion under strong compression can produce locally fluent text while losing substantial sequence-specific content.

This observation suggests that the main difficulty is not the number of latents alone, but the amount of structure that must be generated jointly from noise.
We therefore generate the latent sequence group by group, so that each diffusion pass only needs to recover a subset of the latent states while conditioning on an already generated prefix.
Specifically, we partition the latent sequence into groups of $G$ consecutive blocks, where each group is
\begin{equation}
\mathcal G_g
=
\{(g-1)G+1,\ldots,gG\},
\end{equation}
and factorize the latent flow prior as
\begin{equation}
p_\theta(z_{1:K})
=
\prod\nolimits_{g=1}^{K/G}
p_\theta
\left(
z_{\mathcal G_g}
\mid
z_{<\mathcal G_g}
\right).
\end{equation}
The $G$ latents within each group are denoised jointly, while previously generated groups provide clean context for the next group.
Grouping therefore replaces one difficult joint diffusion problem over the complete latent sequence with a sequence of smaller conditional diffusion problems.
It also makes latent generation incremental: rather than denoising the entire latent sequence at once, the model extends the sequence one group at a time, following a generation pattern similar to block-wise diffusion in discrete DLMs~\citep{arriola2025block}.

The group size $G$ determines how much parallelism is retained.
When $G=K$, all block latents are generated jointly, recovering fully parallel latent diffusion.
When $G=1$, the latent sequence is generated one block at a time, giving each prediction the largest amount of preceding context.
Intermediate values retain parallel generation within each group while reducing the number of unknown latent states that must be recovered jointly.
We use $G=8$ in the main model and study the effect of the group size in our experiments.

\textbf{Conditional group diffusion.}
For each group, we model the next $G$ block latents conditioned on the generated latent prefix.
Consider group $\mathcal G_g$ beginning at block $s=(g-1)G+1$, with committed prefix $z_{<s}=(z_1,\ldots,z_{s-1})$.
We use a block-causal Transformer $B_\theta(\cdot)$ to encode the committed prefix and produce target-specific conditioning states for the group.
Each target position $i\in\mathcal G_g$ is represented by a learned positional query $u_i$ that attends to the complete clean prefix, yielding
\begin{equation}
c^{(g)}
=
c_{s:s+G-1}
=
B_\theta
\left(
z_{<s},\,u_{\mathcal G_g}
\right).
\end{equation}
Conditioned on these representations, we jointly generate the $G$ target latents with rectified flow~\citep{liu2023flow,lipman2023flow}.
For a clean target group $z^{(g)}\in\mathbb{R}^{G\times d}$, we construct
\begin{equation}
z_t^{(g)}
=
t z^{(g)}
+
(1-t)\epsilon,
\qquad
\epsilon\sim\mathcal N(0,I),
\qquad
t\in[0,1].
\end{equation}
A Transformer denoiser $f_\theta$ predicts the clean target group as
\begin{equation}
\hat z^{(g)}
=
f_\theta
\left(
z_t^{(g)},t,c^{(g)}
\right).
\end{equation}

Following ELF~\citep{hu2026elf}, we directly predict the clean latent $z^{(g)}$ and optimize the induced velocity-matching objective,
\begin{equation}
\mathcal L_{\mathrm{flow}}
=
\mathbb E
\left[
\left\|
\frac{\hat z^{(g)}-z_t^{(g)}}{\max(1-t,\epsilon_t)}
-
\frac{z^{(g)}-z_t^{(g)}}{\max(1-t,\epsilon_t)}
\right\|_2^2
\right],
\end{equation}
where $\epsilon_t>0$ is a small constant that prevents numerical instability near the clean endpoint $t=1$.

\textbf{Grouped sampling.}
At inference, BLD generates the latent sequence one group at a time.
For each group, the model computes its conditioning states from the committed latent prefix, initializes the next $G$ latents from Gaussian noise, and denoises them for $T$ integration steps.
The resulting group is then committed as clean context for the next group.
Detailed sampling pseudocode is provided in Appendix~\ref{app:grouped_sampling}.

The prior therefore requires $K/G$ sequential group-generation passes.
For $K=64$ and $G=8$, this gives eight passes, with eight latents denoised jointly per pass.
At $G=K$, all $K$ latents are denoised jointly without a committed prefix, recovering fully parallel latent diffusion.
Our fully parallel reference uses this formulation with a separately tuned training and sampling recipe described in Appendix~\ref{app:parallel_prior}.
Thus, $G$ controls the trade-off between latent-level parallelism and conditional generation difficulty.

\subsection{Branching Token Realization}

Once the latents $z_{1:K}$ have been generated, the branching decoder generates the corresponding token spans in parallel.
Let $x_{i,1:m_i}$ denote the token span associated with $z_i$.
The decoder factorizes token generation as
\begin{equation}
p_\psi(x_{1:L}\mid z_{1:K})
=
\prod\nolimits_{i=1}^{K}
\prod\nolimits_{j=1}^{m_i}
p_\psi
\left(
x_{i,j}\mid x_{i,<j},z_{1:K}
\right).
\end{equation}
Each branch is therefore autoregressive only within its own span and does not condition on tokens generated by other branches.

For branch $i$, a causal Transformer models the local token history $x_{i,<j}$ while cross-attending to the latents $z_{1:K}$.
The local latent $z_i$ provides span-specific information, while the remaining latents provide cross-block context.
This shared context allows independently decoded branches to remain consistent without exchanging generated tokens.

A length head predicts the span length,
\begin{equation}
m_i=\arg\max \ell_\psi(z_i).
\end{equation}
All $K$ branches are decoded in parallel, reducing the sequential token-decoding depth from $L$ to $\max_i m_i$.
For approximately balanced blocks, $\max_i m_i\approx L/K$, giving about 16 sequential steps for $L=1024$ and $K=64$.


Inference proceeds in two stages.
First, the latent prior generates the complete latent sequence in $K/G$ sequential groups, using $T$ denoising steps per group.
Once all latents are available, the decoder predicts the span lengths $m_{1:K}$ and generates the token spans in parallel, with autoregression confined to each span.
The sequential computation therefore consists of $(K/G)T$ latent denoising steps followed by $\max_i m_i$ token-decoding steps, approximately $L/K$ for balanced blocks.

\subsection{Training}

We train BLD in separate stages to decouple representation learning, token generation, and latent generation.

\textbf{Block autoencoding.}
We first jointly train the fusion encoder and branching decoder to reconstruct the input sequence from the compressed latent representation.
The reconstruction objective is
\begin{equation}
\mathcal L_{\mathrm{rec}}
=
-\log
p_\psi
\left(
x_{1:L}\mid q_\phi(x_{1:L})
\right),
\end{equation}
and we additionally optimize a cross-entropy objective for block-length prediction.

To encourage information sharing across blocks, we apply contiguous latent-block masking during autoencoder training.
Specifically, we replace contiguous spans of block latents with learned mask states and require the decoder to reconstruct the corresponding token spans from the remaining latent context.
This discourages individual block latents from encoding only local information and encourages information relevant to each span to be distributed across the latent sequence.
After this stage, the encoder is frozen and defines the latent space modeled by the prior.

\textbf{Decoder adaptation.}
The autoencoding decoder is trained only on encoder latents, whereas at inference it receives latents generated by the prior.
To reduce this train--test mismatch, we further train the decoder on a mixture of clean encoder latents and latent predictions obtained by denoising noisy encoder latents with a trained prior.
The decoder loss is not back-propagated through the prior.
In our experiments, the decoder is adapted once using the fully parallel prior and then shared across all latent priors (Section~\ref{sec:dec_adapt} and Appendix~\ref{app:parallel_prior}).

\textbf{Grouped prior training.}
After freezing the encoder, we train the grouped prior on standardized latent sequences.
For each group, the target latents are corrupted with Gaussian noise, while preceding groups remain clean and serve as conditioning context.
Because the complete clean latent sequence is available during training, all groups can be trained in parallel with teacher forcing despite being generated autoregressively at inference.
Causal masking prevents each group from accessing its own or future clean latents.
We optimize the prior using $\mathcal L_{\mathrm{flow}}$, while keeping the encoder and decoder frozen.
Detailed training pseudocode is provided in Appendix~\ref{app:training_algorithm}.

\section{Experiments}
\label{sec:exp}

We evaluate whether latent compression can reduce the cost of diffusion language generation without substantially degrading generation quality.
We first compare BLD with continuous and discrete DLMs and an autoregressive baseline in terms of generation quality, computational cost, and end-to-end runtime.
We then study how the latent group size controls the trade-off between parallel generation and access to clean conditioning context.
Diagnostic analyses of why strong compression makes latent generation difficult are summarized in Appendix~\ref{sec:insights}. 
Additional block-length measurements and a comparison with Cosmos appear in Appendix~\ref{app:additional_results}.

\subsection{Experimental Setup}

\textbf{Data and model scale.}
Our BLD and AR models are trained on OpenWebText using the T5 tokenizer ($|V|\approx 32$k), with the last 10{,}000 source documents held out for evaluation.
We use sequences of $L=1024$ tokens, each compressed into $K=64$ block latents of dimension $d=512$, corresponding to a token-to-latent compression ratio of $r=16$.
The fusion encoder contains 36M parameters, and the 12-layer branching decoder contains 83M parameters.
Both components are shared across all latent priors, including the fully parallel variant described below.
The grouped prior with $G=8$ contains 511M parameters.
The fully parallel variant ($G=K$) instead denoises all 64 block latents jointly with a single 684M bidirectional denoiser.
The fully parallel model uses a separately optimized training and sampling recipe described in Appendix~\ref{app:parallel_prior}.
Parameter counts in Table~\ref{tab:main} include only components used at generation time, i.e., the prior and the branching decoder ($511\text{M}+83\text{M}=594\text{M}$ for $G=8$ and $684\text{M}+83\text{M}=767\text{M}$ for $G=K$).
All grouped priors are trained for 20k steps with 512 sequences per step, corresponding to approximately $10.5$B training tokens.
We use a learning rate of $3\times10^{-4}$ with cosine decay and random-offset sequence crops.

\textbf{Baselines.}
We compare with ELF~\citep{hu2026elf}, the closest token-level continuous DLM baseline.
We also evaluate the released OpenWebText checkpoints of the discrete masked diffusion model MDLM~\citep{sahoo2024simple} and the block diffusion model BD3-LM~\citep{arriola2025block}, with block sizes $L'\in\{4,8,16\}$.
These models use the GPT-2 tokenizer and were trained with a substantially larger token budget than BLD; their comparisons are therefore not training-budget matched.
We evaluate the released ELF-B, ELF-M, and ELF-L checkpoints (105M, 342M, and 652M parameters) using their default sampling settings: 32-step SDE for ELF-B and 64-step SDE for ELF-M and ELF-L, all with self-conditioning CFG scale 3.
Computation and runtime, which ELF does not report, are measured by us on the released checkpoints with the same settings.
Because ELF differs from BLD in tokenizer and training budget, the quality comparison should be read as indicative rather than fully controlled.
The AR model is trained on the same data and token budget as BLD and is sampled using nucleus sampling with $p=0.95$.
We also include an OpenWebText reference consisting of 1{,}000 training sequences.

\textbf{Evaluation.}
For Gen-PPL and entropy, we follow the evaluation setup of ELF~\citep{hu2026elf}.
Each model generates 1{,}000 unconditional sequences of 1024 tokens.
(i) For generation quality, we report generation perplexity (\textbf{Gen-PPL}) of the full sequence under GPT-2 Large.
Because repetitive text lowers Gen-PPL, we report the average per-sequence unigram entropy (\textbf{Entropy}) of the GPT-2 tokens.
We also report \textbf{MAUVE}~\citep{pillutla2021mauve} between generated and held-out sequences.
For the OpenWebText reference, Gen-PPL and entropy are computed on the 1{,}000 training sequences, while MAUVE is computed against 1{,}000 held-out sequences.
(ii) For generation computation, we report the total \textbf{TFLOPs} of the generation pipeline, computed with the PyTorch flop counter.
We separately report single-sequence latency at batch size 1 and generation throughput at batch size 32.
Runtime comparisons are made only between models evaluated on the same GPU.
Unless otherwise specified, grouped BLD models use 50 SDE sampling steps.

\subsection{Main Results}

\begin{table*}[t]
\centering
\small
\caption{
Generation quality and computation for approximately 1024-token sequences.
Params counts the parameters used at generation time (prior and branching decoder for BLD; the fusion encoder is not used during generation).
States is the number of positions processed by the denoiser.
ELF values use the released checkpoints and their default sampling settings; TFLOPs are measured by us.
$^\dagger$Fully parallel variant with a separately tuned recipe (Appendix~\ref{app:parallel_prior}).
}
\label{tab:main}
\setlength{\tabcolsep}{7pt}
\begin{tabular}{lrrrrrr}
\toprule
Model & Params & States & Gen-PPL $\downarrow$ & Entropy $\uparrow$ & MAUVE $\uparrow$ & TFLOPs $\downarrow$ \\
\midrule
OpenWebText  & -- & -- & 15.3 & 5.32 & 0.95 & -- \\
\midrule
AR & 730M & 1024 & 26.5 & 5.37 & 0.92 & 1.6 \\
\midrule
MDLM & 170M & 1024 & 35.2 & 5.30 & 0.79 & 298.4 \\
BD3-LM, $L'=4$ & 170M & 1024 & 23.4 & 5.27 & 0.82 & 17.9 \\
BD3-LM, $L'=8$ & 170M & 1024 & 28.7 & 5.31 & 0.86 & 17.3 \\
BD3-LM, $L'=16$ & 170M & 1024 & 32.4 & 5.33 & 0.85 & 18.7 \\
\midrule
ELF-B, 32 steps & 105M & 1024 & 23.8 & 5.15 & 0.80 & 7.2 \\
ELF-M, 64 steps & 342M & 1024 & 22.2 & 5.18 & 0.93 & 51.2 \\
ELF-L, 64 steps & 652M & 1024 & 23.9 & 5.29 & 0.91 & 96.4 \\
\midrule
BLD, $G=8$ & 594M & 64 & 26.6 & 5.17 & 0.82 & 1.1 \\
BLD, $G=8$, 32 steps & 594M & 64 & 25.3 & 5.14 & 0.76 & 0.8 \\
BLD, $G=K$$^\dagger$ & 767M & 64 & 24.6 & 5.08 & 0.77 & 4.8 \\
\bottomrule
\end{tabular}
\end{table*}


\begin{table}[t]
\centering
\small
\caption{
End-to-end runtime on a single NVIDIA H100 80GB GPU.
Latency is the median over five post-warm-up runs at batch size 1, and throughput is measured at batch size 32.
Both include sampling and token decoding.
All models use eager execution at the listed precision; details are provided in Appendix~\ref{app:runtime}.
$^\dagger$Fully parallel variant (Appendix~\ref{app:parallel_prior}).
}
\label{tab:runtime}
\setlength{\tabcolsep}{6pt}
\begin{tabular}{llrr}
\toprule
Model & Precision & Latency (s) $\downarrow$ & Throughput (doc/s) $\uparrow$ \\
\midrule
MDLM & bf16 & 30.15 & 0.4 \\
BD3-LM, $L'=4$ & fp32 & 16.77 & 0.9 \\
BD3-LM, $L'=8$ & fp32 & 15.37 & 1.0 \\
BD3-LM, $L'=16$ & fp32 & 14.89 & 1.0 \\
\midrule
ELF-B, 32 steps & bf16 & 0.46 & 18.2 \\
ELF-B, 64 steps & bf16 & 0.90 & 9.5 \\
ELF-M, 64 steps & bf16 & 1.73 & 3.0 \\
ELF-L, 64 steps & bf16 & 2.25 & 1.4 \\
\midrule
AR & fp32 & 12.57 & 1.5 \\
\midrule
BLD, $G=8$ & fp32 & 2.67 & 9.4 \\
BLD, $G=8$, 32 steps & fp32 & 1.84 & 13.4 \\
BLD, $G=K$$^\dagger$ & fp32 & 1.16 & 13.5 \\
\bottomrule
\end{tabular}
\end{table}

Tables~\ref{tab:main} and~\ref{tab:runtime} compare generation quality, computation, and end-to-end runtime.

\textbf{Generation quality.}
Despite compressing 1024 tokens into only 64 latent states, BLD retains competitive generation quality.
At $G=8$, BLD achieves a Gen-PPL of 26.6 and entropy of 5.17, closely matching the AR baseline in perplexity and operating at an entropy comparable to ELF-B and ELF-M.
Its MAUVE score of 0.82 exceeds ELF-B at 0.80, but remains below ELF-M, ELF-L, and AR at 0.93, 0.91, and 0.92, respectively.
The fully parallel $G=K$ variant lowers Gen-PPL to 24.6, but also reduces entropy to 5.08 and MAUVE to 0.77.
Thus, its lower perplexity does not translate into a better overall quality--diversity trade-off.
Overall, BLD preserves competitive generation quality under $16\times$ latent compression, although a distributional gap to the strongest baselines remains.

\textbf{Generation cost.}
Latent compression substantially reduces computation.
BLD with $G=8$ requires only 1.1 TFLOPs per sequence, compared with 7.2 TFLOPs for ELF-B, 51.2 TFLOPs for ELF-M, and 96.4 TFLOPs for ELF-L.
Relative to the similarly sized ELF-L model, this corresponds to an $88\times$ reduction in generation FLOPs.
BLD also requires less computation than the AR baseline at 1.6 TFLOPs.
The fully parallel $G=K$ variant requires 4.8 TFLOPs, whereas grouped generation with $G=8$ reduces this to 1.1 TFLOPs.
These results demonstrate the computational benefit of performing diffusion over 64 compressed latent states rather than 1024 token-level states.

\textbf{End-to-end runtime.}
The FLOP reduction translates clearly into higher batched throughput, while single-sequence latency remains affected by sequential group-generation passes.
On the H100, BLD with $G=8$ reaches 9.4 sequences per second, compared with 3.0 for ELF-M, 1.4 for the similarly sized ELF-L, and 1.5 for AR.
This corresponds to a $6.7\times$ throughput improvement over ELF-L and a $6.3\times$ improvement over AR.
At batch size 1, BLD takes 2.67 seconds, slightly longer than ELF-L at 2.25 seconds but substantially shorter than AR at 12.57 seconds.
ELF-B remains faster in single-sequence latency because it uses a much smaller network and only 32--64 denoising steps.

The fully parallel $G=K$ variant reduces latency to 1.16 seconds and reaches 13.5 sequences per second, but at substantially higher computational cost than $G=8$.
The grouped model therefore trades some single-sequence latency for lower FLOPs while retaining high batched throughput.

\subsection{Effect of Group Size}
\label{sec:group}

The group size $G$ determines how many latent states are generated jointly at each stage.
Smaller groups expose more previously generated latents as clean context, but require more sequential generation stages.
Larger groups increase latent-level parallelism at the cost of jointly resolving more uncertainty within each group.

\begin{table}[t]
\centering
\small
\caption{
Effect of the group size $G$.
All models use the same compressed latent representation, branching decoder, grouped prior architecture, and training recipe; only $G$ varies.
Latency is measured at batch size 1 on an H200.
}
\label{tab:group}
\setlength{\tabcolsep}{6pt}
\begin{tabular}{lrrrr}
\toprule
$G$ & Gen-PPL $\downarrow$ & Entropy $\uparrow$ & Latency (s) $\downarrow$ & TFLOPs $\downarrow$ \\
\midrule
1 & 31.7 & 5.22 & 4.93 & 2.23 \\
4 & 24.7 & 5.21 & 3.52 & 1.31 \\
8 & 26.6 & 5.17 & 1.84 & 1.10 \\
16 & 27.7 & 5.03 & 1.21 & 1.00 \\
\bottomrule
\end{tabular}
\end{table}

Table~\ref{tab:group} shows the trade-off induced by the group size.
Increasing $G$ from 1 to 4 substantially improves Gen-PPL while leaving entropy nearly unchanged.
Increasing $G$ further to 8 and 16 progressively reduces latency and computation, but Gen-PPL begins to increase and entropy decreases.
Latency falls from 4.93 seconds at $G=1$ to 1.21 seconds at $G=16$, while computation decreases from 2.23 to 1.00 TFLOPs.

We use $G=8$ as the default operating point.
Compared with $G=4$, it roughly halves latency and reduces computation from 1.31 to 1.10 TFLOPs, with a modest increase in Gen-PPL from 24.7 to 26.6.
Although $G=16$ is faster still, it further increases Gen-PPL to 27.7 and lowers entropy to 5.03.
Somewhat unexpectedly, $G=1$ yields the worst Gen-PPL even though each prediction has the largest clean prefix.
We attribute this to exposure: with $G=1$, generation involves 64 sequential commitments, so errors in self-generated context compound over more stages (Appendix~\ref{app:exposure}).

We examine how clean prefix information is exposed to the latent states within each group.
With shared group-level conditioning, all target latents receive the same summary of the generated prefix.
This becomes increasingly restrictive for larger groups and can lead to repetition or collapse near group boundaries.
Target-specific conditioning instead provides each target latent with its representation of the clean prefix.
This largely removes the boundary failure mode at $G=8$ without increasing inference cost, raising distinctness from approximately 0.22 to 0.33 and reducing the hard-loop rate from 14.5\% to 4.3\% (Appendix~\ref{app:conditional_structure}); we use target-specific conditioning in the main model.

\subsection{Effect of Decoder Adaptation}
\label{sec:dec_adapt}

The decoder in the unadapted control is trained on clean and diffusion-noised encoder latents, but has never seen a prior's outputs.
We therefore adapt the decoder to denoised latent predictions before generation.
To isolate this effect, we fix the $G=8$ prior and vary only the decoder.

\begin{table}[t]
\centering
\small
\caption{
Effect of decoder adaptation with the same $G=8$ prior.
Rec.\ denotes reconstruction NLL from exact encoder latents.
}
\label{tab:dec_adapt}
\setlength{\tabcolsep}{5pt}
\begin{tabular}{lrrrr}
\toprule
Decoder & Gen-PPL $\downarrow$ & Entropy $\uparrow$ & MAUVE $\uparrow$ & Rec.\ $\downarrow$ \\
\midrule
Not adapted & 127 & 5.34 & 0.67 & 0.042 \\
Adapted on $G=K$ & 26.6 & 5.17 & 0.82 & 0.030 \\
Adapted on $G=K$, then $G=8$ & 27.1 & 5.15 & 0.78 & 0.027 \\
\bottomrule
\end{tabular}
\end{table}

Table~\ref{tab:dec_adapt} shows that decoder adaptation substantially improves generation with the fixed $G=8$ prior: adapting on $G=K$ reduces Gen-PPL from 127 to 26.6 and increases MAUVE from 0.67 to 0.82. Further adaptation on $G=8$ slightly improves reconstruction NLL but does not improve generation quality. These results suggest that the main benefit comes from exposing the decoder to prior-generated latents, and that this adaptation transfers well across priors without requiring specialization to the evaluation prior.

\section{Related Work}

\textbf{Continuous diffusion language models.}
Continuous DLMs generate text by denoising continuous representations~\citep{li2022diffusion,lin2023text,gao2024empowering,shabalin2025tencdm,hu2026elf}.
Most retain approximately one state per token, so the sequence processed by the denoiser scales with the output length.
BLD instead denoises a compressed sequence of block-level latents.

\textbf{Discrete and block diffusion language models.}
Discrete DLMs iteratively unmask or resample tokens~\citep{lou2024discrete,sahoo2024simple,nie2025large}, while block diffusion~\citep{arriola2025block} generates token blocks autoregressively with parallel denoising within each block.
These methods retain one state per token.
BLD adopts a similar block-autoregressive pattern in compressed continuous space, denoising $G$ latents that jointly represent about $rG$ tokens per step.

\textbf{Latent and compressed diffusion for language.}
Prior work explores language diffusion in learned latent spaces~\citep{lovelace2023latent,zhang2023planner}.
Most closely related, Cosmos~\citep{meshchaninov2025cosmos} learns compressed, noise-robust representations and shows that reconstruction quality alone does not determine diffusion quality.
Its strongest results focus on shorter sequences and more moderate compression.
BLD targets $16\times$ compression over 1024 tokens through block-structured latents, grouped generation, and \emph{branching token realization}.

\textbf{Block- and patch-level language modeling.}
MEGABYTE~\citep{yu2023megabyte} and BLT~\citep{pagnoni2025blt} separate global patch-level modeling from local token prediction, while CALM~\citep{shao2025calm} autoregressively predicts continuous representations encoding multiple tokens.
SSD-LM~\citep{han2023ssdlm} introduces blockwise progression into diffusion-based generation.
BLD combines compressed latent denoising with branching token realization, reducing the denoising sequence length while preserving parallel token generation.
\section{Conclusion}
\label{sec:conclusion}

We presented \emph{Branching Latent Diffusion (BLD)}, a framework that moves continuous language diffusion from token-level states to compressed block latents.
BLD combines compressed latent generation with \emph{branching token realization}, allowing local AR branches to decode token spans in parallel.
To make generation under strong compression more tractable, BLD generates the latent sequence in groups conditioned on previously generated latents.
Experiments show that this design substantially reduces generation FLOPs and improves end-to-end batch throughput relative to size-matched token-level continuous DLM and AR baselines, and reduces single-sequence latency relative to the AR baseline.
Long-range coherence nevertheless remains a significant challenge, indicating that efficient compressed generation still requires better modeling of global semantic structure.
Overall, our results suggest that moving diffusion to shorter latent sequences is a promising direction for efficient long-sequence language generation.

\textbf{Limitations.}
(i) \emph{Quality and scope.}
BLD performs worse than the strongest token-level continuous DLMs on MAUVE and Gen-PPL.
We evaluate only unconditional generation on OpenWebText using Gen-PPL, entropy, and MAUVE, without human evaluation or a direct measure of long-range coherence.
(ii) \emph{Comparisons.}
ELF and the discrete diffusion baselines differ from BLD in tokenizer and training budget, and the fully parallel variant uses a different training recipe. We lack training-matched comparisons with these models and a length-matched comparison with compressed latent diffusion.
(iii) \emph{Latency.}
Grouped sampling requires $(K/G)T$ sequential denoiser evaluations; its batch-throughput gains do not establish a single-sequence latency advantage.
(iv) \emph{Scale.}
We study one model scale, sequence length ($L=1024$), and full-system compression ratio ($r=16$), leaving their effects on the quality--efficiency trade-off open.

\clearpage
\subsection*{AI use statement}

We used generative AI tools to edit and polish the manuscript for clarity, readability, and language quality.
During experimentation, we also used generative AI tools to assist in writing and debugging auxiliary scripts used for development and troubleshooting.
We did not use generative AI tools to formulate the core research ideas, design the main methodology, or interpret the experimental results.
All AI-assisted outputs were reviewed and verified by the authors, who take full responsibility for the final content of this work.

\bibliography{dlm}
\bibliographystyle{iclr2027_conference}

\clearpage
\appendix

\startcontents[appendix]
\section*{Appendix}
\printcontents[appendix]{}{1}[2]{}

\newpage

\section{Additional Analysis of Compressed Continuous Diffusion}
\label{app:analysis}

\subsection{Understanding Compressed Latent Diffusion}
\label{sec:insights}

Our analysis examines how strong compression changes the latent generation problem beyond its computational cost.
We summarize three empirical findings below and provide supporting experiments in the following subsections.

\textbf{Insight 1: Reconstruction quality does not determine diffusion quality.}
Compressed latents can retain enough information for accurate reconstruction while becoming substantially harder to recover from noise.
Across both controlled compression experiments and learned representations, better reconstruction does not consistently translate into better generation.
Representation quality must therefore be assessed together with the difficulty of modeling the resulting latent distribution.

\textbf{Insight 2: Ambiguous conditioning can produce poorly decodable predictions.}
Under squared-error denoising, the optimal clean-state prediction is a conditional mean, which may lie between valid latent states.
Our interpolation and conditional probes show that such intermediate predictions can decode poorly even when the decoder tolerates local perturbations around real latents.
These observations motivate reducing denoising ambiguity through informative conditioning.

\textbf{Insight 3: Target-specific access to clean context improves grouped generation.}
Generating latents in groups provides each group with previously generated latents as clean conditioning context.
How this context is exposed also matters: replacing a shared prefix summary with target-specific conditioning substantially improves generation at $G=8$.
These findings motivate combining grouped latent generation with position-specific access to the committed prefix.

\subsection{Reconstructability and Generatability}
\label{app:reconstructability}

A compressed representation must preserve the information needed for token realization, but reconstruction quality alone does not predict how reliably a diffusion prior can generate it.

\paragraph{Controlled compression ladder.}
We isolate the effect of compression in an ELF-B-based continuous diffusion model by varying the number of tokens represented by each latent state.
After the frozen token encoder, $n$ adjacent token latents are mean-pooled into one state; a position-conditioned readout predicts the $n$ tokens from that state.
The Transformer architecture, optimizer, number of updates, data, and sampler are shared across arms, while the latent scale is re-estimated for each $n$ and the denoising compute changes with sequence length.
All variants are trained for five epochs (95{,}085 updates) with a global batch size of 512.
Table~\ref{tab:compression_ladder} compares clean-latent reconstruction with recovery of the source text from the midpoint of the diffusion trajectory.

\begin{table}[htbp]
\centering
\small
\caption{
Token accuracy under increasing latent compression.
Clean reconstruction decodes the uncorrupted latent.
Midpoint recovery measures recovery of the source text from a latent corrupted to the midpoint of the diffusion trajectory.
Both columns report token accuracy on 64 validation sequences using the final five-epoch checkpoint; midpoint recovery decodes a one-step clean-state prediction at $t_0=0.5$ with self-conditioning guidance scale 1.
}
\label{tab:compression_ladder}
\begin{tabular}{lrr}
\toprule
Tokens per latent & Clean reconstruction & Midpoint recovery \\
\midrule
1 & 1.000 & 0.925 \\
2 & 0.999 & 0.791 \\
4 & 0.952 & 0.455 \\
\bottomrule
\end{tabular}
\end{table}

At two tokens per latent, clean reconstruction is essentially unchanged, while recovery from noise is substantially worse.
At four tokens per latent, both degrade, but the reduction in midpoint recovery is much larger.
These results show that denoising difficulty can increase before clean reconstruction deteriorates substantially.
Source recovery is a diagnostic of denoising behavior, however, and should not be equated with unconditional generation quality.

At eight and sixteen tokens per latent, the simple pooling interface no longer reconstructs the text reliably.
These settings therefore confound representation loss with denoising difficulty, whereas the two- and four-token settings more clearly separate the two effects.

\paragraph{Fixed dimensionality reduction can improve generation despite losing information.}
We observe a related trade-off in the block-latent representation.
Starting from a $512$-dimensional latent, we project each block onto nested PCA subspaces and train a fresh prior in the retained coordinates.
The PCA basis is fitted once on block latents from 4{,}096 training documents; the subspaces are nested and no retained coordinate is rescaled.
Each rank uses the same frozen autoencoder and a fresh development-stage prior with four learned planning slots (Plan-4), which provide global conditioning to the block-latent denoiser, trained for 5{,}000 steps.
Table~\ref{tab:pca_rate} reports clean decoding from projected real latents (oracle PPL) and decoding from generated latents under paired sampling noise.

\begin{table}[t]
\centering
\small
\caption{
PCA rate scan on a frozen $K=64$, $L=1024$ block-latent autoencoder.
Each prior is trained for 5{,}000 steps; generation uses 256 paired samples, greedy decoding, and a 256-token GPT-2 scoring window.
Oracle PPL scores decoded real latents after projection; generated PPL scores prior samples.
The transport gap is $\log(\mathrm{generated\ PPL}/\mathrm{oracle\ PPL})$.
These development-stage scores are not directly comparable with Table~\ref{tab:main}.
}
\label{tab:pca_rate}
\begin{tabular}{rrrrr}
\toprule
Rank & Retained variance & Oracle PPL & Generated PPL & Transport gap \\
\midrule
512 & 1.000 & 30.56 & 554.19 & 2.898 \\
256 & 0.879 & 31.84 & 441.01 & 2.628 \\
224 & 0.834 & 36.34 & 476.70 & 2.574 \\
192 & 0.780 & 55.66 & 393.98 & 1.957 \\
128 & 0.635 & 199.38 & 259.11 & 0.262 \\
\bottomrule
\end{tabular}
\end{table}

At rank 192, oracle PPL rises from 30.56 to 55.66 while generated PPL falls from 554.19 to 393.98; distinctness also rises from 0.466 to 0.476.

An ambient-dimensional control retaining the same PCA subspace produces nearly the same result.
This suggests that the improvement comes primarily from changing the retained information rather than reducing the denoiser's input dimensionality.
Reducing the rank further to $128$ makes generation easier relative to the oracle decoding floor, but discards too much information for useful reconstruction.
In this experiment, rank $192$ provides a better balance between representation distortion and latent generation error.

\paragraph{Learned bottlenecks can reverse this improvement.}
We next replace the fixed PCA projection with a learned $192$--$224$ dimensional bottleneck.
These representations rapidly recover near-lossless oracle decoding, but generation deteriorates as autoencoder training continues.
For example, at autoencoder steps 5k, 10k, and 15k, the learned $192$-dimensional code has oracle PPL 33.68, 31.08, and 30.77, while corresponding fresh priors produce generated PPL 2201.84, 2850.24, and 3165.75.
Each prior is freshly trained for the corresponding autoencoder checkpoint using the same development-stage Plan-4 protocol as in the fixed-PCA experiment.
The fixed PCA-$192$ representation remains substantially easier to generate despite its worse oracle decoding floor.

A stochastic autoencoder exhibits similar behavior.
Training the decoder to reconstruct from a noisy $192$-dimensional channel improves reconstruction and local robustness, yet fresh priors trained on these latents generate substantially worse text than those trained on the fixed PCA representation.

These results suggest that nominal dimensionality and local decoder robustness are insufficient measures of generative difficulty.
A reconstruction-trained encoder can place information in joint dependencies that the decoder reads accurately but the tested prior does not reliably reproduce.
The relevant trade-off therefore involves both reconstruction distortion and the difficulty of generating the encoded information.

\subsection{Conditional Ambiguity under Compression}
\label{app:conditional_ambiguity}

We next examine how the prior and decoder behave when a conditioning state is compatible with multiple clean realizations.

\paragraph{Intermediate document representations can decode poorly.}
We interpolate between the latent representations of two held-out documents while keeping the encoder and decoder fixed.
The probe uses 64 held-out documents and no further training.
Table~\ref{tab:latent_interpolation} shows that small perturbations toward another document are largely harmless, whereas the midpoint matches neither parent and decodes into incoherent text.

\begin{table}[t]
\centering
\small
\caption{
Decoding interpolated real-document latents $z(a)=(1-a)z_A+az_B$ with a frozen encoder and decoder.
Exact match measures agreement with document $A$; PPL is GPT-2-scored on a 256-token window.
The probe uses 64 held-out documents and is separate from the main generation evaluation.
}
\label{tab:latent_interpolation}
\begin{tabular}{rrr}
\toprule
$a$ & Exact match with $A$ & PPL \\
\midrule
0.125 & 99.9\% & 30.6 \\
0.250 & 97.8\% & 33.6 \\
0.375 & 67.5\% & 141.6 \\
0.500 & 19.0\% & 680.8 \\
\bottomrule
\end{tabular}
\end{table}

At $a=0.5$, the PPL is close to the corresponding prior's free-running value of 682 under its development-stage protocol.

Block transplantation provides a complementary control.
Replacing complete latent blocks with those from another document causes the edited spans to decode as the donor document while leaving the remaining blocks nearly unchanged.
Across transplants of 1, 4, or 16 contiguous blocks, donor blocks decode at 99.9--100\% exact match and untouched blocks at 99.94\%.
The decoder can therefore handle locally valid but globally mixed latent assignments.
The interpolation failure appears more closely associated with intermediate states within individual blocks than with mixing blocks from different documents.

\paragraph{Ambiguous conditions can induce averaged predictions.}
A memorization experiment probes the prior side of this behavior.
We train a model to recover the latent residual associated with one of a finite set of memorized document plans.
This diagnostic model is trained on only 1{,}024 documents and evaluated with raw weights and a 50-step ODE sampler.
Adding normalized plan noise with standard deviation 0.05--0.4 changes paired residual cosine only from 0.9951 to 0.9938 and PPL from 31.38 to 31.54; nearest-mode retrieval remains 100\%.
At the midpoint between two memorized plans, cosine to the two parents falls to 0.541/0.536, detail standard deviation falls to 0.767, and PPL rises to 226.9.
An unseen-plan control gives cosine 0.0273 and PPL 395.71, so the near-exact recovery demonstrates memorization rather than generalization to new documents.

Together, these probes suggest a possible failure mechanism.
Conditions that identify a particular realization support accurate predictions, whereas ambiguous conditions can produce intermediate predictions that the decoder interprets poorly.
This concerns the behavior of individual denoising predictions; it does not imply that complete diffusion trajectories must terminate at conditional means.

\paragraph{Additional denoising can sharpen predictions without recovering the source.}
Starting from noisy versions of real latents, we compare one-step clean predictions with the results of continuing the full sampler.
Additional sampling often reduces decoder uncertainty, but does not consistently improve token accuracy relative to the source document.
At moderate and high noise levels, the final sample can become sharper while source accuracy decreases.
At low noise levels, additional sampling can reduce both source fidelity and generation quality.

Recovering the particular source document is unnecessary for unconditional generation.
These observations instead show that sharper predictions and better source recovery are distinct outcomes.
Additional denoising may select a plausible realization without recovering the one used to construct the noisy input.

Changes to the number of sampling steps, stochasticity, noise scale, and terminal integration behavior often move generation along a quality--diversity frontier in our experiments.
Such changes affect the sampling trajectory, but do not provide additional observed context.
This motivates examining conditioning structure alongside sampler design.

\subsection{The Role of Conditional Structure}
\label{app:conditional_structure}

The same compressed representation can behave differently depending on how much of the latent sequence is available as context and how each target accesses it.
Here, \emph{clean} denotes latents provided without additional diffusion noise.
Such latents may come from the encoder during training or from previously generated groups during inference.

\paragraph{Clean prefixes improve the use of cross-block information.}
Fully parallel generation denoises the complete latent sequence while every block is uncertain.
Although the denoiser can attend across blocks, neighboring states are themselves noisy.
In this setting, generated latents retain substantially less cross-block structure than real latent sequences.

We compare this behavior with a block-autoregressive prior trained on the same frozen latent representation.
Conditioning each new block on a clean prefix produces stronger cross-block dependence and richer document-specific content.
This suggests that the representation retains cross-block information that is more effectively used under clean-prefix conditioning.

\begin{table}[t]
\centering
\small
\caption{
Development-stage block-autoregressive (AR) latent prior and fully parallel diffusion control on the same frozen autoencoder and decoder.
Gen-PPL is scored by GPT-2 Small on 256-token windows from 256 samples, rather than the full-document protocol of Table~\ref{tab:main}.
The AR prior uses a 400M backbone and 53M flow head with 50 SDE steps per block; the parallel prior has 684M parameters and uses 50 steps over the whole sequence.
}
\label{tab:ar_latent_diagnostic}
\begin{tabular}{lrrr}
\toprule
Prior & Gen-PPL & Entropy & Distinct-1 \\
\midrule
Block AR, noise scale 0.6 & 19.3 & 5.35 & 0.252 \\
Fully parallel, noise scale 0.7 & 28.1 & 5.30 & 0.220 \\
\bottomrule
\end{tabular}
\end{table}

The block-AR advantage in Table~\ref{tab:ar_latent_diagnostic} is a short-window diagnostic, not a full-length quality claim.
In inspected 1024-token AR samples, coherence deteriorates after roughly token 512 and tails enter name, citation, or byline patterns.
The archived analysis traces this position-linked shift partly to the training corpus packing, which lacks random-offset cropping.

\paragraph{Group size controls how much structure is generated jointly.}
Grouped generation interpolates between sequential and fully parallel latent generation.
At each stage, the clean prefix is fixed and the next $G$ latent blocks are generated jointly.
Increasing $G$ reduces the number of sequential stages, but increases the amount of unresolved structure within each group.

An initial grouped prior used a single prefix representation to condition all target blocks.
This design remained usable for small groups but became unstable for larger groups.
At $G=8$ and $G=16$, generation frequently collapsed near group boundaries into repeated phrases, lists, or other low-diversity patterns.
The training loss changed only modestly, indicating that these failures were poorly reflected by the flow objective.

\paragraph{Target-specific conditioning improves group-boundary behavior.}
We replace the shared conditioning vector with one query per target block.
Each query occupies its target block's position and attends directly to the complete clean prefix.
The latent representation, decoder, group size, and training budget are otherwise unchanged.

At $G=8$, this modification largely removes the immediate group-boundary collapse.
Distinctness increases from approximately $0.22$ to $0.33$, while the hard-loop rate decreases from approximately $14.5\%$ to $4.3\%$.
The segment-level diversity profile also shows less degradation after the first group boundary.
Here, distinctness is the mean, over 256 generated sequences, of the fraction of unique T5 tokens in each full sequence.
The hard-loop rate is the fraction of those sequences whose first 256 tokens contain an identical 8-gram at least three times.

These results distinguish the availability of context from its accessibility to each target.
A clean prefix provides conditioning information, while the conditioning architecture determines how that information reaches individual target states.
Target-specific access substantially improves generation in this setting.

\paragraph{More direct prefix access changes the remaining failures.}
We test two additional conditioning architectures.
In one variant, the group denoising head cross-attends to the full clean prefix at every layer and denoising step.
In another, the clean prefix and noisy target group are processed jointly within a block-causal Transformer.

Both variants improve the retention of prefix information.
Names and subject-specific fragments persist further into generated continuations, and the joint Transformer improves teacher-forced regression error relative to the separate-head baseline.
However, more direct access can also produce copied names, phrase permutations, and repeated local fragments.
The joint Transformer produces cleaner unconditional documents than the cross-attention head, but still exhibits register transitions and incoherent tails.

These observations suggest that access to context and effective use of context are separate challenges.
The denoiser must use document-specific information to generate a compatible continuation without simply copying local fragments.

\subsection{Exposure and Remaining Failure Modes}
\label{app:exposure}

Conditional latent generation also faces a mismatch between real context during training and generated context during inference.
The following diagnostics examine this mismatch and its interaction with conditioning.

\paragraph{Real prefixes expose both conditional ability and its limits.}
Under teacher forcing, the target group is conditioned on exact encoder latents.
At inference, the same model receives previously sampled groups.
When 32 or 48 real blocks are clamped as a prefix, the $G=8$ prior often retains the prefix's register and domain for two or three sampled groups, but loses a named subject in the first generated group.
In the eight inspected continuations, rare price-list and chat-log formats also collapse immediately.
Thus, generated-prefix exposure contributes to later drift, while a real prefix alone does not resolve the initial loss of document-specific content.

This difference also limits anti-repetition objectives evaluated only on real prefixes.
Repeated states encountered during generation may be uncommon in teacher-forced training, leaving such objectives with little direct signal about those failures.
Including model-generated context during training directly probes this mismatch, but does not by itself establish a remedy.

\paragraph{One-step prefix rollout has limited benefits for grouped priors.}
In diagnostic rollout-training experiments, selected real context blocks are replaced with detached model predictions.
At $G=4$, this delays the onset of list and byline patterns and reduces list-driven entity counts, but six of eight inspected documents still enter such a pattern.
At $G=1$, fluency worsens; at $G=8$ and $G=16$, all inspected documents still loop or enter lists near the first group boundary.
The one-step replacement therefore does not reproduce all the errors encountered with fully sampled prefixes.

These observations separate context-distribution mismatch from the weakness of the conditional itself: the latter is visible even at the first sampled block after a real prefix.
The reported grouped priors do not use rollout training; the fully parallel reference does, as described in Appendix~\ref{app:parallel_prior}.

\paragraph{Near-clean context rollout improves the fully parallel prior but leaves loops.}
We separately fine-tuned the earlier 684M fully parallel prior for 15k steps while replacing eligible near-clean context blocks with detached model predictions, then re-noising those predictions before conditioning other blocks.
The original fine-tuning run serves as the control; R1 replaces eligible blocks with probability 0.5, R3 increases this to 1.0, and R4 also lowers the eligibility threshold from $t_{\min}=0.85$ to 0.7.
Table~\ref{tab:parallel_rollout} gives representative sampling points, which differ in entropy and should be read with the within-campaign quality--diversity comparisons rather than ranked by PPL alone.

\begin{table}[t]
\centering
\small
\caption{
Development-stage rollout fine-tuning of the fully parallel prior, with 50-step sampling.
Gen-PPL is scored by GPT-2 Large on 256-token windows; loop rate is measured on 50 inspected full texts per point.
Noise scales and resulting entropies differ across rows; these scores are not the full-document scores in Table~\ref{tab:main}.
}
\label{tab:parallel_rollout}
\begin{tabular}{lrrrrr}
\toprule
Fine-tuning & Noise scale & Gen-PPL & Entropy & Distinct-1 & Loop rate \\
\midrule
Control & 0.70 & 25.83 & 5.304 & 0.220 & 0.86 \\
R1 ($p=0.5$) & 0.60 & 26.77 & 5.43 & 0.228 & 0.82 \\
R3 ($p=1$) & 0.60 & 24.65 & 5.395 & 0.220 & 0.90 \\
R4 ($p=1$, $t_{\min}=0.7$) & 0.70 & 23.10 & 5.195 & 0.210 & -- \\
\bottomrule
\end{tabular}
\end{table}

Across two seeds, R3 improves PPL by approximately four points at entropy near 5.41 and by about 1.3 points at matched distinctness, but its loop rate remains comparable to the control.
R4's lower PPL accompanies lower entropy and distinctness; at similar PPL, the R3 setting retains higher entropy.
Thus, rollout training improves conditional robustness in this fully parallel recipe without resolving its repetition attractor.

\paragraph{Decoder adaptation has limited ability to repair latent errors.}
The branching decoder reconstructs high-quality text from exact encoder latents and tolerates modest latent errors after adaptation.
However, different decoder variants exhibit similar semantic failures when given the same degraded latent sequence.
Improving clean-latent reconstruction can also increase sensitivity to details that the prior does not reliably reproduce.

These results indicate that cross-block semantic consistency depends strongly on the generated latent sequence.
Decoder adaptation can absorb some latent distortion, but does not reliably repair missing or inconsistent document-level information.
BLD therefore assigns global structure to the latent sequence while restricting token-level autoregression to individual branches.

\section{Algorithm Details}

\subsection{Grouped Sampling Algorithm}
\label{app:grouped_sampling}

Algorithm~\ref{alg:grouped-generation} generates each group conditioned on the committed latent prefix.
The backbone computes the target-specific conditioning states once per group, and the flow head reuses them throughout denoising.

\begin{algorithm}[t]
\caption{BLD Grouped Latent Generation}
\label{alg:grouped-generation}
\begin{algorithmic}[1]
\Require Number of blocks $K$, group size $G$, sampling steps $T$
\Require Sampling grid $\{t_n\}_{n=0}^{T}$ with $t_0=0$, noise scale $\sigma$, churn rate $\gamma$, denominator floor $\epsilon_t$
\Require Block-causal backbone $B_\theta$, target queries $u_{1:K}$, flow head $f_\theta$
\Ensure Generated latent sequence $z_{1:K}$

\State $z_{1:K}\gets\varnothing$

\For{$g=1,\ldots,K/G$}
    \State $s\gets(g-1)G+1$
    \State $\mathcal G_g\gets\{s,\ldots,s+G-1\}$
    \State $c^{(g)}\gets B_\theta(z_{<s},u_{\mathcal G_g})$
    \State $z^{(g)}\gets\sigma\epsilon$,
    \quad $\epsilon\sim\mathcal N(0,I)$

    \For{$n=0,\ldots,T-1$}
        \State $\tilde t_n\gets t_n$

        \If{stochastic sampling}
            \State $\alpha\gets\operatorname{clip}\bigl(1-\gamma(t_{n+1}-t_n),0,1\bigr)$
            \State $\tilde t_n\gets\alpha t_n$
            \State $z^{(g)}\gets
            \alpha z^{(g)}
            +(1-\alpha)\sigma\epsilon'$,
            \quad $\epsilon'\sim\mathcal N(0,I)$
        \EndIf

        \State $\hat z^{(g)}
        \gets
        f_\theta(z^{(g)},\tilde t_n,c^{(g)})$

        \State $v
        \gets
        \dfrac{\hat z^{(g)}-z^{(g)}}
        {\max(1-\tilde t_n,\epsilon_t)}$

        \State $z^{(g)}
        \gets
        z^{(g)}
        +(t_{n+1}-\tilde t_n)v$
    \EndFor

    \State Commit $z_{\mathcal G_g}\gets z^{(g)}$
\EndFor

\State \Return $z_{1:K}$
\end{algorithmic}
\end{algorithm}

For the reported $G=8$ sampler, $T=50$, $t_n=n/T$, $\gamma=1$, $\sigma=0.7$, and $\epsilon_t=0.05$; the final integration time is $t_T=1$.
The stochastic update, including its $(1-\alpha)\sigma$ noise coefficient, follows the implementation in \texttt{ARLatentPrior.\_sample\_group}.
It is an SDE-style re-noising heuristic; this coefficient is not chosen to preserve the marginal variance of the linear interpolation at $\tilde t_n$.

\subsection{Grouped Latent Prior Training}
\label{app:training_algorithm}

Algorithm~\ref{alg:grouped-training} summarizes training of the grouped prior.
Training uses clean encoder latents as teacher-forced context, allowing all conditioning states to be computed in one block-causal backbone pass.
The flow objectives for different groups are then evaluated in parallel.

\begin{algorithm}[t]
\caption{Grouped Latent Prior Training}
\label{alg:grouped-training}
\begin{algorithmic}[1]

\Require Training sequence $x_{1:L}$, number of blocks $K$, latent dimension $d$, group size $G$
\Require Time distribution $p(t)$, denominator floor $\epsilon_t$
\Require Frozen encoder $q_\phi$, block-causal backbone $B_\theta$, target queries $u_{1:K}$, flow head $f_\theta$
\Ensure Flow-matching loss $\mathcal L_{\mathrm{flow}}$

\State $z_{1:K}
\gets
\operatorname{Normalize}\left(q_\phi(x_{1:L})\right)$

\State Partition $z_{1:K}$ into groups
$\mathcal G_1,\ldots,\mathcal G_{K/G}$,
with $z^{(g)}=z_{\mathcal G_g}$

\State Compute conditioning states
$c^{(1)},\ldots,c^{(K/G)}$
with one block-causal backbone pass

\ForAll{$g=1,\ldots,K/G$ \textbf{in parallel}}
    \State Sample $t_g\sim p(t)$ and
    $\epsilon^{(g)}\sim\mathcal N(0,I)$

    \State $z_{t_g}^{(g)}
    \gets
    t_g z^{(g)}
    +(1-t_g)\epsilon^{(g)}$

    \State $\hat z^{(g)}
    \gets
    f_\theta
    \left(
    z_{t_g}^{(g)},t_g,c^{(g)}
    \right)$

    \State $\hat v^{(g)}
    \gets
    \dfrac{
    \hat z^{(g)}-z_{t_g}^{(g)}
    }{
    \max(1-t_g,\epsilon_t)
    }$

    \State $v^{(g)}
    \gets
    \dfrac{
    z^{(g)}-z_{t_g}^{(g)}
    }{
    \max(1-t_g,\epsilon_t)
    }$
\EndFor

\State $\mathcal L_{\mathrm{flow}}
\gets
\dfrac{1}{Kd}
\sum_{g=1}^{K/G}
\left\|
\hat v^{(g)}-v^{(g)}
\right\|_F^2$

\State Update the trainable prior parameters using $\mathcal L_{\mathrm{flow}}$

\State \Return $\mathcal L_{\mathrm{flow}}$

\end{algorithmic}
\end{algorithm}

The attention mask preserves the conditional factorization used during generation.
Queries for group $\mathcal G_g$ can access clean blocks preceding that group, but cannot access its target blocks or future blocks.
The model therefore learns conditionals of the form
\begin{equation*}
p_\theta
\left(
z_{\mathcal G_g}
\mid
z_{<\mathcal G_g}
\right)
\end{equation*}
without processing the groups sequentially during optimization.
Sequential group generation is required only at inference, when sampled groups replace the real teacher-forced prefixes.

\section{Implementation Details}

\subsection{Fully Parallel Latent Prior}
\label{app:parallel_prior}

The fully parallel reference ($G=K$) in Table~\ref{tab:main} is a separately developed prior rather than the grouped architecture with a single group.
It jointly denoises all $K=64$ block latents using a bidirectional Transformer with hidden size 1536, 24 layers, 24 heads, and 684M parameters.
It uses the same frozen fusion encoder as the grouped priors.
The decoder adapted on this parallel prior is also shared with the grouped models.

\paragraph{Choice of fully parallel reference.}
With a single group, the grouped architecture has no committed prefix.
Its backbone conditioning states are therefore independent of the sampled sequence, leaving the 117M flow head to perform sample-dependent denoising.
In our experiments, a fully parallel prior of about 100M parameters collapsed onto a few repeated topics even with the training recipe below.
We therefore report the strongest fully parallel prior we developed.

This prior was tuned for longer than the grouped priors and uses a different training and sampling recipe.
The differences are detailed below.

\paragraph{Per-block noise levels.}
In half of the training batches, each block receives an independent noise level, following diffusion forcing~\citep{chen2024diffusion}.
The remaining batches use one shared noise level.
For a subset of training sequences, some blocks are kept clean as context and the loss is evaluated only on the remaining blocks.

\paragraph{Training schedule.}
The prior is trained for 30k steps at a learning rate of $3\times10^{-4}$ and then fine-tuned for 15k steps at $10^{-4}$ with cosine decay.
During fine-tuning, near-clean context blocks are replaced with model predictions for rollout training, as discussed in Appendix~\ref{app:exposure}.
The branching decoder is trained jointly on clean latents and predictions denoised by this prior.

The first stage uses 64 sequences per step, corresponding to approximately 2B tokens.

\paragraph{Sampling.}
All $K$ latents are denoised with a 50-step SDE sampler and a noise scale of $0.6$.
The blocks follow staggered noise schedules in a random order drawn for each sample.
Blocks that finish earlier provide partially resolved context for the remaining blocks.

\paragraph{Scope of the comparison.}
The grouped priors use neither per-block noise levels nor rollout training and process approximately 10.5B training tokens; the fully parallel reference processes approximately 2B tokens in its first stage, with its fine-tuning token budget still to be confirmed.
The adapted decoder is shared, so decoder adaptation is not a difference between the reported models.
The comparison in Table~\ref{tab:main} therefore evaluates the strongest configurations developed for the two designs rather than isolating the effect of grouping.
Table~\ref{tab:group} provides the controlled comparison across group sizes within the grouped architecture.

\subsection{Inference Stack for Runtime Measurements}
\label{app:runtime}

All runtimes are measured on one NVIDIA H100 80GB HBM3 GPU with no other job on the device.
Every model runs in eager PyTorch without \texttt{torch.compile} or CUDA graphs; ELF's \texttt{use\_compile} option is disabled.
Latency is the median of five batch-1 runs after warm-up, and throughput is measured on one batch of 32 documents.
Timings use synchronized wall-clock measurements from initial noise, or the first token for AR, to output token IDs.
They include sampling and decoding but exclude model loading and detokenization.

The samplers use PyTorch's \texttt{scaled\_dot\_product\_attention}, with the backend selected at runtime.
Profiler inspection of the BLD, AR, and ELF runs shows that fp32 uses the memory-efficient kernel, bf16 uses FlashAttention-2 and cuDNN attention, and ELF's masked attention uses cuDNN attention.
BLD and the token AR baseline run with PyTorch 2.12, CUDA 13.0, and Transformers 5.10.
ELF runs in its released environment with PyTorch 2.14 and Transformers 4.44.

\paragraph{BLD.}
The grouped prior samples the $K/G$ groups sequentially.
For each group, the backbone encodes the committed prefix once without a key/value cache across groups, requiring $K/G$ backbone passes.
The flow head then runs 50 SDE steps with $\gamma=1$, for $50K/G$ head passes in total.
The fully parallel reference runs 50 SDE steps with $\gamma=1.4$ over all 64 latents, using the staggered schedule described in Appendix~\ref{app:parallel_prior}.

The branching decoder generates all 64 blocks in parallel with incremental decoding.
It maintains per-layer key/value caches for within-block self-attention and computes cross-attention keys and values over the latent sequence once.
Token selection is greedy, and each branch stops at its predicted length.
We report fp32 measurements because bf16 does not improve performance in this inference stack (Table~\ref{tab:runtime_precision}).
The samplers issue many small kernels, while autocast introduces additional cast operations.

\paragraph{AR.}
The AR baseline uses Hugging Face \texttt{generate} with a key/value cache, SDPA attention, nucleus sampling with $p=0.95$, and a fixed generation length of 1024 tokens.
We report fp32 measurements, which are faster than bf16 in this stack.
For the 730M model, runtime is measured using randomly initialized weights with the same architecture as the trained model: 24 layers, width 1536, and a 32k vocabulary.

\paragraph{ELF.}
We use the released sampler, \texttt{\_generate\_samples\_single\_batch} followed by \texttt{\_dlm\_decode\_batch}, with the README settings.
SDE sampling uses $\gamma=1.5$ for 32 steps and $\gamma=1.0$ for 64 steps.
Self-conditioning CFG uses scale 3.
Sampling uses bf16 autocast through \texttt{use\_bf16}.
Every denoising step processes the complete 1024-position sequence without a key/value cache.

\paragraph{MDLM and BD3-LM.}
We use the released OpenWebText checkpoints and the BD3-LM repository's first-hitting sampler with nucleus threshold $p=0.9$.
Each 1024-token document takes 1023 token-reveal calls; BD3-LM uses a key/value cache, while MDLM processes the full sequence at every call.
We retain the first draw of every sample rather than applying the repository's low-entropy rejection rule.
Batch-1 latency uses the repository sampler; batch-32 throughput uses a batched implementation of the same reveal procedure, which samples only at the position selected for reveal rather than drawing candidates at all masked positions.
MDLM uses bf16 autocast and BD3-LM uses fp32 computation with a bf16 key/value cache, matching their released inference settings.
These runs use PyTorch 2.7.1 and Transformers 4.49.

\begin{table}[htbp]
\centering
\small
\caption{
Runtime under different precisions, using the same GPU and measurement protocol as Table~\ref{tab:runtime}.
Entries report latency (s) / throughput (doc/s).
For bf16 weights, parameters are cast to bf16 once and inference also uses bf16 autocast.
}
\label{tab:runtime_precision}
\begin{tabular}{lrrr}
\toprule
Model & fp32 & bf16 autocast & bf16 weights \\
\midrule
BLD, $G=8$ & 2.67 / 9.4 & 3.76 / 5.8 & 3.46 / 5.9 \\
BLD, $G=K$ & 1.16 / 13.5 & 1.40 / 13.2 & 1.33 / 13.4 \\
AR & 12.57 / 1.5 & 16.15 / 0.56 & 14.29 / 0.58 \\
\bottomrule
\end{tabular}
\end{table}

\section{Additional Results}
\label{app:additional_results}

\subsection{Block and Branch Lengths}
\label{app:block_lengths}

We analyze encoder block lengths and generation-time branch lengths for BLD ($L=1024$, $K=64$).
The shared fusion encoder selects the highest-scoring boundary within each window around the uniform grid, allowing spans of 1--31 tokens.
For documents with $n<L$ tokens, the windows shrink to approximately $n/K$ positions.
At generation time, branch lengths are predicted by the decoder's length head.
Table~\ref{tab:block_lengths} reports encoder spans on training and held-out documents, and predicted branch lengths for 1000 unconditional samples from each BLD model, using the samplers in Table~\ref{tab:main}.

\begin{table}[htbp]
\centering
\small
\caption{
Encoder block lengths and predicted decoder branch lengths ($L=1024$, $K=64$).
Full length: documents with at least $L$ tokens.
All lengths: held-out documents with at least $K$ tokens, truncated to $L$ and right-padded as in training (median length 983; 2.5\% reach $L$).
Longest span denotes $\max_i m_i$ per document.
Parentheses give boundary-position base rates over all token positions.
}
\label{tab:block_lengths}
\resizebox{\linewidth}{!}{%
\begin{tabular}{lrcrrcrr}
\toprule
& & & \multicolumn{2}{c}{share of blocks} & longest span & \multicolumn{2}{c}{boundary position (\%)} \\
\cmidrule(lr){4-5}\cmidrule(lr){7-8}
Documents & docs & $m_i$ (mean $\pm$ std) & $m_i=16$ & $m_i\le4$ & median / p95 / max & word start & after sentence end \\
\midrule
\multicolumn{8}{l}{\emph{Encoder (chunker), real documents}} \\
Held-out, full length & 330 & $16.00\pm0.18$ & 96.7\% & 0.0\% & 17 / 17 / 17 & 65.8 (65.3) & 3.9 (3.7) \\
Training, full length & 1000 & $16.00\pm0.31$ & 96.5\% & $<$0.1\% & 17 / 17 / 31 & 64.8 (64.8) & 3.6 (3.7) \\
Held-out, all lengths & 1000 & $14.98\pm3.17$ & 84.9\% & 6.0\% & 16 / 22 / 24 & 69.0 (69.2) & 3.8 (3.6) \\
\midrule
\multicolumn{8}{l}{\emph{Decoder (length head), generated documents}} \\
BLD, $G=8$ & 1000 & $15.17\pm2.91$ & 88.0\% & 4.6\% & 16 / 22 / 24 & -- & -- \\
BLD, $G=K$ & 1000 & $15.96\pm0.54$ & 96.8\% & 0.1\% & 16 / 17 / 24 & -- & -- \\
\bottomrule
\end{tabular}}
\end{table}

\paragraph{Encoder boundaries are nearly uniform.}
On full-length documents, 99.3\% of internal boundaries fall one token before their grid point, typically yielding a 15-token first block, 16-token interior blocks, and a 17-token last block.
Boundary scores strongly favor positions $t\equiv15\pmod{16}$, while alignment with word starts and sentence endings remains close to the base rates.
For shorter documents, narrower windows may contain no preferred position, causing the 16-token intervals to split into shorter spans.

\paragraph{Dynamic versus uniform chunking.}
The near-uniform inference boundaries do not establish that learning the boundaries is immaterial during training.
A development-stage paired experiment changed only the chunker: dynamic and strictly uniform variants shared the data order, seed, and 20k-step autoencoder recipe at $L=1024$, $K=64$.
Both then used matched 45k-step prior training and a six-layer decoder.
Table~\ref{tab:chunking_twin} reports their latent spectra and generation diagnostics.

\begin{table}[t]
\centering
\small
\caption{
Paired dynamic-versus-uniform chunking diagnostic on a reduced development stack.
PR and $k_{99}$ characterize the autoencoder latent spectrum; reconstruction CE uses real latents.
Generation points have approximately matched distinctness, but their tested entropy ranges do not overlap, so the PPL entries are not an iso-entropy comparison and must not be compared with Table~\ref{tab:main}.
}
\label{tab:chunking_twin}
\resizebox{\linewidth}{!}{%
\begin{tabular}{lrrrrrrr}
\toprule
Chunker & PR & $k_{99}$ & Recon. CE & Gen-PPL & Entropy & Distinct-1 & Loop rate \\
\midrule
Dynamic & 9.7 & 180 & 0.156 & 105.4 & 5.831 & $\approx$0.305 & 0.12 \\
Uniform & 33.3 & 243 & 0.041 & 343.2 & 6.049 & $\approx$0.305 & 0.38 \\
\bottomrule
\end{tabular}}
\end{table}

Uniform chunking reconstructs real latents more accurately but yields a broader latent spectrum and worse generated text in this paired setup.
This result supports a training-time contribution from dynamic chunking without identifying the precise mechanism or measuring its effect on the final $G=8$ prior.
In a separate single-variable probe on a uniform grid, replacing mean pooling with learned attention pooling left $k_{99}$ essentially unchanged (392 versus 393); that probe stopped at the representation gate and did not evaluate generation.

\paragraph{Generated branch lengths and decoding depth.}
Branch lengths under $G=8$ resemble encoder spans across all document lengths, whereas those under $G=K$ resemble spans on full-length documents.
The two priors generate 971 and 1021 tokens per document on average, respectively.
The token-decoding depth $\max_i m_i$ has a median of 16 for both priors, with 95th percentiles of 22 and 17 for $G=8$ and $G=K$, respectively, and an observed maximum of 24 for both.
Thus, $L/K=16$ approximates the typical decoding depth, although longer branches increase it for some documents.

\subsection{Selected Unconditional Samples}
\label{app:samples}

We show two contiguous passages selected from the unconditional outputs of the main-table BLD checkpoints.

\paragraph{BLD, $G=8$.}
\begin{Verbatim}[fontsize=\footnotesize]
The programs have added new opportunities for growth, helping local
investors to develop new markets for consumers. In addition, cheaper
options for buying and selling goods to attract visitors and investors.
The latest expansion in the region has been criticized for developing
an efficient relationship with emerging and tourism markets. The
company also offers to local residents. The new partnership will
provide more opportunities for integrating local industry and
infrastructure to help sustain and maintain sustainable production.
\end{Verbatim}

\paragraph{BLD, $G=K$.}
\begin{Verbatim}[fontsize=\footnotesize]
The Canadian government says it's a bestselling blog for
Business-DuffPost.com—a $500,000 market for biotechnology companies.
It's not just a multimillion-dollar business to grow GMO products,
which are used to produce products for potential customers. The
business of bio-based products is a major factor requiring consumers
to buy products for high-end products. They are aiming to sell
products for foreign-owned companies to the world's consumers.
\end{Verbatim}

\subsection{Comparison with Cosmos}
\label{app:cosmos}

Cosmos~\citep{meshchaninov2025cosmos} is the closest related latent diffusion language model.
Its released OpenWebText model generates at most 512 BERT tokens using 512 latent vectors of dimension 768.
We sample 1000 documents from the released checkpoints (autoencoder: 200k steps; diffusion: 500k steps) using the official sampler.
Sampling uses EMA weights, 200 Euler--Maruyama steps from $t=1$ to $t=0.05$, self-conditioning, and argmax decoding terminated at the first \texttt{[SEP]}.

The outputs have a median length of 473 GPT-2 tokens, with 93\% shorter than 512.
We evaluate prefixes of at most $\ell\in\{128,256\}$ GPT-2 tokens using 256 generated documents and the same 1000 reference documents as Table~\ref{tab:main}, truncated identically.
At these lengths, 5.1\% and 9.0\% of Cosmos outputs remain shorter than $\ell$, respectively.
Table~\ref{tab:cosmos} reports these measurements alongside the main-text results for context.
Because evaluation lengths differ, the quality scores do not provide a controlled ranking of the models.

\begin{table}[htbp]
\centering
\small
\caption{
Cosmos measurements and main-text results under different evaluation lengths.
Baseline and BLD scores use approximately 1024-token sequences; Cosmos scores use prefixes of at most 128 or 256 GPT-2 tokens.
Quality scores across these settings are not directly comparable.
Length denotes maximum output length in each model's tokenizer; states denotes generator sequence positions.
TFLOPs reports cost per full generated document before truncation.
$^\dagger$Fully parallel reference (Appendix~\ref{app:parallel_prior}).
}
\label{tab:cosmos}
\resizebox{\linewidth}{!}{%
\begin{tabular}{lrrrrrr}
\toprule
Model & Length & States & Gen-PPL $\downarrow$ & Entropy $\uparrow$ & MAUVE $\uparrow$ & TFLOPs $\downarrow$ \\
\midrule
\multicolumn{7}{l}{\emph{Main-text evaluation: approximately 1024 tokens}} \\
OpenWebText & -- & -- & 15.3 & 5.32 & 0.95 & -- \\
AR & 1024 & 1024 & 26.5 & 5.37 & 0.92 & 1.6 \\
ELF-B, 32 steps & 1024 & 1024 & 23.8 & 5.15 & 0.80 & 7.2 \\
BLD, $G=8$ & 1024 & 64 & 26.6 & 5.17 & 0.82 & 1.1 \\
BLD, $G=K$${}^\dagger$ & 1024 & 64 & 24.6 & 5.08 & 0.77 & 4.8 \\
\midrule
\multicolumn{7}{l}{\emph{Cosmos evaluation: short GPT-2 prefixes}} \\
Cosmos, $\ell=128$ & 512 & 512 & 106.0 & 4.14 & 0.926 & 26.7 \\
Cosmos, $\ell=256$ & 512 & 512 & 84.1 & 4.51 & 0.939 & 26.7 \\
\bottomrule
\end{tabular}}
\end{table}

Cosmos costs 26.7 TFLOPs per full generated document, approximately $24\times$ the cost of BLD with $G=8$.
Using the Cosmos evaluation protocol with regex tokenization replacing spaCy, we obtain perplexity 98.9 and diversity 0.346, compared with the reported 55.0 and 0.319.
The perplexity discrepancy remains unresolved.
All Cosmos results reported here are our own measurements.

\end{document}